\documentclass[runningheads]{llncs}
\usepackage{graphicx}

\usepackage{tikz}
\usepackage{comment}
\usepackage{amsmath,amssymb} 
\usepackage{color}

\usepackage{graphicx}
\usepackage{amsmath}
\usepackage{amssymb}
\usepackage{booktabs}
\usepackage{microtype}
\usepackage{units}
\usepackage{bbold}
\usepackage{tabularx}
\usepackage{multirow}
\usepackage{pifont}
\newcommand{\cmark}{\ding{51}}%
\newcommand{\xmark}{\ding{55}}%
\newcommand{\etal}{\emph{et al}.}

\usepackage[accsupp]{axessibility}  
\usepackage[pagebackref,breaklinks,colorlinks]{hyperref}

\begin{document}
\pagestyle{headings}
\mainmatter
\def\ECCVSubNumber{6827}  
\title{Embedding contrastive unsupervised features to cluster in- and out-of-distribution noise in corrupted image datasets}

\titlerunning{Embedding contrastive unsupervised features to cluster OOD noise}
\author{Paul Albert, 
Eric Arazo, Noel E. O'Connor, Kevin McGuinness
{\tt\small paul.albert@insight-centre.org}
\institute{School of Electronic Engineering, \\
Insight SFI Centre for Data Analytics, Dublin City University (DCU)}}
\authorrunning{Albert~\etal} 

\maketitle

\begin{abstract}
   Using search engines for web image retrieval is a  tempting alternative to manual curation when creating an image dataset, but their main drawback remains the proportion of incorrect (noisy) samples retrieved. These noisy samples have been evidenced by previous works to be a mixture of in-distribution (ID) samples, assigned to the incorrect category but presenting similar visual semantics to other classes in the dataset, and out-of-distribution (OOD) images, which share no semantic correlation with any category from the dataset. The latter are, in practice, the dominant type of noisy images retrieved. To tackle this noise duality, we propose a two stage algorithm starting with a detection step where we use unsupervised contrastive feature learning to represent images in a feature space. We find that the alignment and uniformity principles of contrastive learning allow OOD samples to be linearly separated from ID samples on the unit hypersphere. We then spectrally embed the unsupervised representations using a fixed neighborhood size and apply an outlier sensitive clustering at the class level to detect the clean and OOD clusters as well as ID noisy outliers. We finally train a noise robust neural network that corrects ID noise to the correct category and utilizes OOD samples in a guided contrastive objective, clustering them to improve low-level features. Our algorithm improves the state-of-the-art results on synthetic noise image datasets as well as real-world web-crawled data. Our work is fully reproducible github.com/PaulAlbert31/SNCF.
\end{abstract}

\begin{figure}[t]
\centering
\includegraphics[width=.4\linewidth]{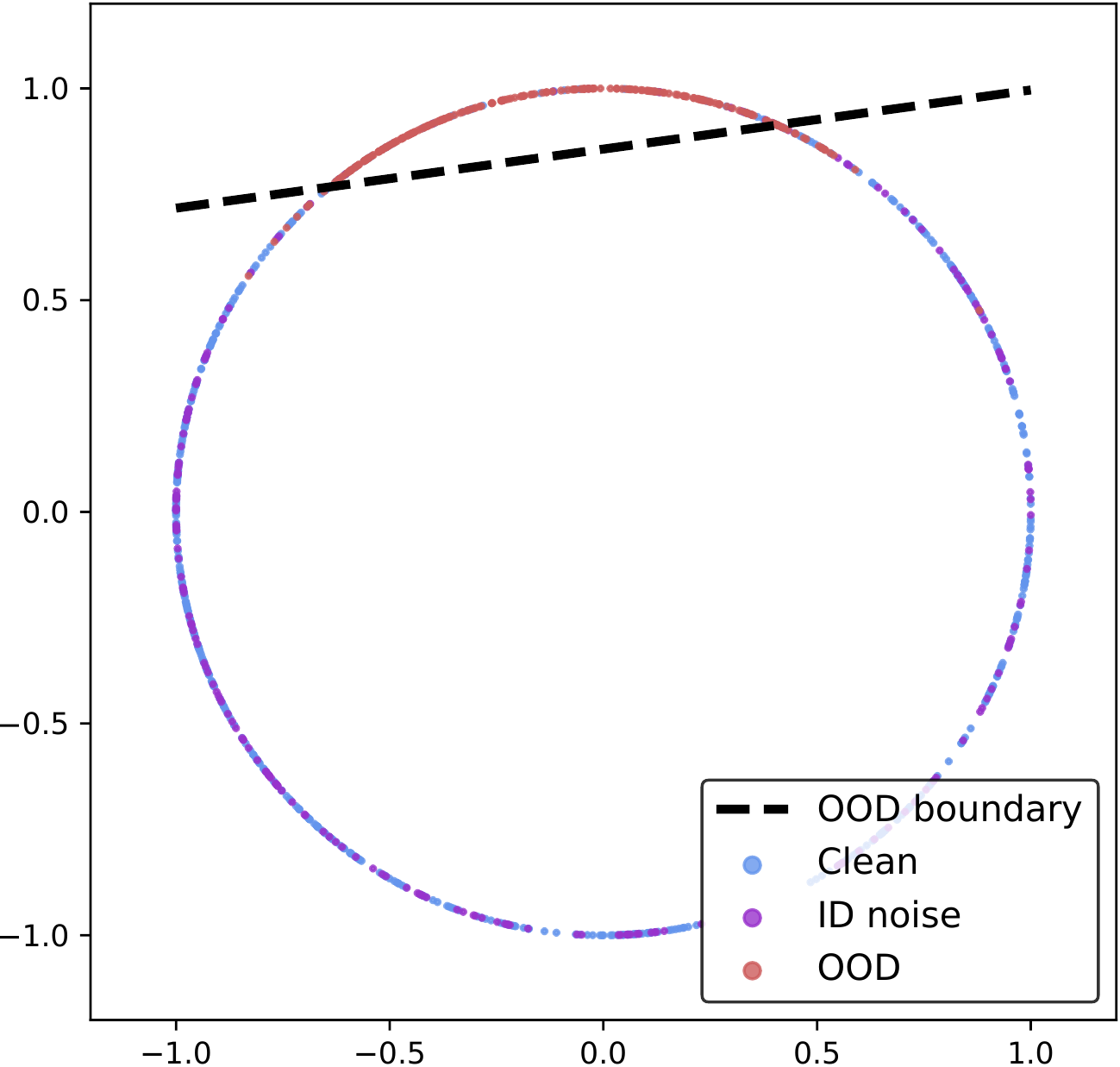}
\par
\caption{Visualization of the linear separation between OOD and ID unsupervised contrastive representations on the 2D hypershpere. CIFAR-10 corrupted with $r_{in} = r_{out} = 0.2$, OOD from ImageNet32. Linear separability in 2D at the dataset level is $92.49\%$ but increases to $98\%+$ for 128D \label{fig:cifar10}}
\end{figure}
\section{Introduction}
Convolutional neural networks (CNN) show remarkable classification accuracies on large image datasets carefully curated by large numbers of annotators~\cite{2016_CVPR_ResNet,2019_ICML_EfficientNet}, which entails that the data gathering and labeling process has become a significant part of the effort required when designing a deep learning image classification algorithm. Recent research proposes to reduce the cost of the data gathering effort in multiple ways, including semi-supervised learning~\cite{2011_AISTATS_STL10} where only a small part of the images have been labeled by humans, which is often the most time consuming part of the dataset building process; unsupervised learning~\cite{2020_Nature_colonunsup}, where visual features are learned without the need for labels; and web-crawled datasets~\cite{2017_arXiv_WebVision} constructed from search engine queries, although this results in the inclusion of incorrect samples assigned to the incorrect class or not belonging to the distribution and corrupting the CNN's convergence. This paper tackles this latter challenge. Designing algorithms capable of training highly accurate CNNs even when trained on imperfect web-crawled data is an important step towards the widespread deployment and take up of computer vision algorithms in practice. CNNs have been shown to completely overfit noisy samples in a dataset without proper regularization~\cite{2017_ICLR_Rethinking}, which degrades performance. More specifically, the noise observed in web-crawled datasets has been categorized as both in-distribution and out-of-distribution, the latter being the dominant type~\cite{2022_WACV_DSOS}. While in-distribution (ID) noisy images can be directly used to train the network after correcting their assigned label, out-of-distribution (OOD) images cannot be assigned to any category. Since a trusted in-distribution dataset is unavailable and the identity of clean and noisy samples is unknown, out-of-distribution detection algorithms~\cite{2019_ICCV_oodpseudounsup,2021_NeurIPS_oodlimit,2019_ICLR_deepanomaly}, which use a classifier trained on clean data to be able to detect OOD samples post training, cannot be used. This further complicates the noise detection process. Once noisy images have been identified, simply ignoring out-of-distribution images has been shown to be sub-optimal as these samples still contain meaningful information for learning low-level features that can be leveraged to improve the representations learned~\cite{2021_ICCV_trashtresure,2021_CVPR_JoSRC}. This paper proposes to tackle the in-distribution and out-of-distribution duality of the noise present in web-crawled datasets specifically to improve the final classification accuracy. To detect the noise, we observe that unsupervised contrastive representations for OOD samples are linearly separated from ID ones on the hypersphere (see Figure~\ref{fig:cifar10}) and train a robust network that will use current representations to correct ID noisy samples and use OOD data to improve low-level representations using contrastive learning. Our contributions are:
\begin{itemize}
    \item A dual noise detection approach utilizing the alignment and uniformity principles of contrastive learning to detect noisy samples in a spectral embedding of unsupervised representations.
    \item A noise robust algorithm capable of training a CNN on a dataset corrupted with in-distribution and out-of-distribution noise, correcting the label of in-distribution whilst using out-of-distribution noise to improve low-level features.
    \item Experiments on controlled and real world noisy datasets demonstrating the state-of-the-art performance of our algorithm.
\end{itemize}

\section{Related work}
\subsection{Web-crawled datasets}
Web-crawled datasets are a low supervision alternative to dataset building that helps to democratize deep learning approaches. Here we will give some examples. Webvision~\cite{2017_arXiv_WebVision} is a dataset constituted of $2.4$ million images gathered using search queries on the same 1k classes as the ILSVRC12~\cite{2012_NeurIPS_ImageNet} challenge. Albert~\etal~\cite{2022_WACV_DSOS} have shown that the noise present in Webvision is predominantly out-of-distribution. Clothing1M~\cite{2015_CVPR_Clothing1M} is a 1M images clothes classification dataset, popular dataset in the label noise community which, as shown by the authors, only contains ID noise. More recently, Sun~\etal~\cite{2021_ICCV_weblyfinegrained} released additional web datasets for fine grained classification tasks.
\subsection{Tackling ID and OOD noise}
Most of the noise robust algorithms that showed improvements on web-noise assume that all the noise is ID. The strategies used to combat ID noise include: robust loss functions~\cite{2015_ICLR_Bootstrapping,2017_CVPR_ForwardLoss}, meta-learning~\cite{2021_CVPR_FaMUS,2020_arXiv_MetaSoftApple}, sample weighing using a mentor network~\cite{2018_ICML_MentorNet,2020_ICML_MentorMix}, network regularization~\cite{2018_ICLR_mixup,2020_NeurIPS_EarlyReg}, semi-supervised learning~\cite{2020_ICLR_DivideMix,2021_ICLR_robustcurri,2020_ICPR_Robust}, contrastive learning to detect the noise in MOIT~\cite{2021_CVPR_MOIT} or improve the label correction using nearest neighbor clustering together with a semi-supervised correction in ScanMix~\cite{2021_arXiv_Scanmix}.
Another class of noise robust algorithms have emerged recently to tackle noisy datasets presenting both in- and out-of-distribution noise.
EvidentialMix~\cite{2020_WACV_EDM} and DSOS~\cite{2022_WACV_DSOS} differentiates between ID noisy and OOD data using a custom noise landscape and JoSRC~\cite{2021_CVPR_JoSRC} evidences OOD samples as having a low agreement between two consistent views of the same image. The methods we compare against are described in more details in~\ref{par:baselines} and we direct the interested reader to a recent label noise survey by Song~\etal~\cite{2020_arXiv_Survey} for an in-depth overview of state-of-the-art label noise robust algorithms.
\section{Algorithm description~\label{sec:description}}
This paper studies image classification in the presence of label noise, where part of the available image dataset $\mathcal{X} = \{x_i\}_{i=1}^N$ and its associated classification labels $\mathcal{Y} = \{y_i\}_{i=1}^N$, with the class distribution $\{c\}_{c=1}^C$, is corrupted by $N_o$ out-of-distribution samples and $N_n$ in-distribution noisy samples, where $N_c = N - N_o - N_n$ is the number of in-distribution clean examples. $N_o$, $N_n$ as well as the identity of the ID noisy and OOD samples are unknown. Examples of such datasets are web-crawled datasets: Webvision~\cite{2017_arXiv_WebVision}, Clothing1M~\cite{2015_CVPR_Clothing1M}, and the Webly Supervised Fine-Grained Recognition datasets~\cite{2021_ICCV_weblyfinegrained}. We propose here an algorithm capable of training a convolutional neural network (CNN) $\Psi$ on the corrupted dataset $\mathcal{X}$ without over-fitting to the noise and capable to accurately classify examples belonging to the class distribution.

\subsection{Unsupervised feature learning}
First, our algorithm learns unsupervised representations from the images themselves, independently of their label. We aim here to relate images to each other in order to capture clusters of similar images. To do so, we train the $N$-pairs unsupervised contrastive learning algorithm which has been successfully used in metric learning~\cite{2016_NeurIPS_NPairs} and unsupervised learning on images and text~\cite{2021_ICLR_iMix}. Given two mini-batches of size $B$ formed from two strongly data augmented views $x_i'$ and $x_i''$ of $x_i \in \mathcal{X}$, we enforce $u_i'$ and $u_i''$, their associated contrastive representations through $\Psi$, to be similar to each other and dissimilar to every other image in the batch. We compute the unsupervised contrastive loss
\begin{equation}
    l_{unsup} = - \frac{1}{B}\sum_{i=1}^{B}\log\left(\frac{\exp{\left(\textit{ip}(u_i'', u_i')/\tau_2\right)}}{\sum_{k=1}^B\exp{\left(\textit{ip}(u_k'', u_i')/\tau_2\right)}}\right),
    \label{eq:unsupcont}
\end{equation}
where $\textit{ip}(u_1,u_2) = \frac{u_1^T.u_2}{\Vert u_1\Vert_2\Vert u_2\Vert_2}$ is the inner product operation, measuring the similarity between contrastive representations, and $\tau_2$ a temperature hyper-parameter, fixed to $0.2$ for every experiment. Mixup~\cite{2018_ICLR_mixup} can be optionally used to further augment $x_i'$ by linearly interpolating it with other augmented samples from the mini-batch with a parameter $\mu$ drawn from a beta distribution with parameter $1$ to produce $x_{mix}' = \mu x_i' + (1-\mu) x_j'$ with $x_j'$ a random sample from the mini-batch (different for every $x_i$) and $u_{mix}'$ the associated representation of $x_{mix}'$. We then use 
\begin{equation}
\begin{aligned}
    l_{mix} = - \frac{1}{B}\sum_{i=1}^{B} & \log\left(\mu\frac{\exp{\left(\textit{ip}(u_i'', u_{mix}')/\tau_2\right)}}{\sum_{k=1}^B\exp{\left(\textit{ip}(u_k'', u_{mix}')/\tau_2\right)}}\right. \\
    & + \left. (1-\mu)\frac{\exp{\left(\textit{ip}(u_j'', u_{mix}')/\tau_2\right)}}{\sum_{k=1}^B\exp{\left(\textit{ip}(u_k'', u_{mix}')/\tau_2\right)}}\vphantom{\frac12}\right),
    \label{eq:unsupcontmixup}
\end{aligned}
\end{equation}
the $N$-pairs loss paired with mixup as a data augmentation. This unsupervised contrastive objective has been proposed as part of the iMix~\cite{2021_ICLR_iMix} algorithm.

\subsection{Embedding of unsupervised features}
We propose not to use the learned unsupervised features directly but instead to perform a non-linear spectral dimensionality reduction on an affinity matrix (embedding). The aim of the embedding is to capture the affinities between samples and their neighbors where ID clean samples will be very similar to other samples from the same class, ID noisy samples will be similar to other ID samples from a different class and OOD samples dissimilar to any ID sample. This motivates computing the embedding at the dataset level to ensure that the similarity of ID noise to other classes is captured.
We first compute the sparse similarity matrix $S$ of size $N\times N$ where for each sample in the dataset, we compute the affinity to a fixed neighborhood size of $50$ neighbors.
\begin{equation}
S_{ij} =\left(\nicefrac{u_{i}^Tu_{j}}{\left\Vert u_{i}\right\Vert_2 \left\Vert u_{j}\right\Vert_2  }\right)^{\gamma},
\end{equation}
with $u_{i}$ the unsupervised representation for sample $x_i$ (not augmented) and $\gamma = 3$ a hyper-parameter regulating the importance of distant neighbors. 
With $I_N$ the identity matrix of size $N$ and $D$ the diagonal normalization matrix where $D_{ii} = \sum_{j=1}^N S_{ij}$, we compute the normalized Laplacian 
\begin{equation}
    L = I_N - D^{-1/2}SD^{-1/2}.
\end{equation}
We finally compute the first $k$ eigenvectors of the normalized Laplacian $L$ by solving
\begin{equation}
    (L - \lambda)V = 0
\end{equation}
and concatenating the first $k$ eigenvectors $V$ of $L$ (by increasing order of the eigenvalues $\lambda$, omitting the smallest), providing us with $k$ features per sample to form the embedding $E$. In practice we use $k=20$ for every dataset. This embedding process is commonly referred to as spectral embedding~\cite{2002_NeurIPS_spectralclust,2000_TPAMI_normalizedcuts}.

\subsection{Unsupervised clustering of noise \label{sec:unsupclust}}
Using the embedding $E$, we cluster embedded unsupervised features to identify three kinds of samples: clean ID, noisy ID, and OOD. In the generic case where the three types of noise are expected in the dataset, we apply the clustering at the class level and aim to discover three clusters for each class: a high density cluster of ID clean samples, a low density cluster of OOD samples and ID noisy outliers. In the case where no ID noise is present, we observe the cluster separation at the dataset level and use a two mode Gaussian mixture to retrieve each cluster.

\textbf{Why does OOD noise cluster?} Contrary to previous research where OOD noise is considered an outlier to the distribution~\cite{2018_CVPR_IterativeNoise}, we observe in this paper that unsupervised contrastive learning can be effectively used to cluster noise in the feature space. We expand here on our intuition as to why this works using the alignment and uniformity principles for contrastive learning formalized by Wang~\etal~\cite{2020_ICLR_understandingcontrastive}. Unsupervised contrastive learning pulls together augmented representations of a same image while pushing apart representations of any other sample in the mini-batch. Since images from a same class will be similar to each other's augmentations, they will cluster together in the feature space to create one (or more) mode for the class (alignment principle). On the other hand, by considering OOD samples as being uniformly sampled from the set of all images, meaning much more varied in appearance than the ID set, we would expect that no compact mode would appear and that these samples would remain uniformly distributed in the feature space yet separated from the ID examples, pulled together into their respective class modes (uniformity principle). Since the features are $L^2$ normalized during training they exist on the surface of a unit hypersphere and one side of the sphere will contain well represented ID classes, clustered into their respective modes, while OOD noise will remain uniformly distributed be on the other side of the hypersphere and linearly separable from ID samples. Section~\ref{sec:linearsep} proposes experiments to evidence the linear separability of ID and OOD samples in the unsupervised contrastive feature space. The spectral embedding we propose has a key role to play in the clustering of the OOD noise which, although separable from the ID samples, is much more spread-out that the compact class modes of ID images. We remedy this problem by computing the affinities in $S$ not over a fixed distance threshold but using a fixed number of neighbors.

\textbf{OPTICS~\cite{1999_ACM_Optics}} is an algorithm which allows us to detect clusters as well as outliers: each feature point is ordered to create a chain where neighboring points are ordered next to each other. Each feature point is then labeled with a reachability cost to neighbors in a neighborhood of size $V$. The higher the cost, the more likely a sample is to be an outlier. Finally, clusters are identified in the ordering where ``valleys'' of low reachability cost evidence a cluster, themselves separated with high cost outliers. 

\textbf{Discovering clean and noisy clusters.} Because the difficulty of learning  similar unsupervised features varies from class to class in an image dataset, we propose to modify the OPTICS algorithm to become more flexible to our problem. We aim here to be able to detect varying valley sizes in the ordered reachability plot where different classes in the image dataset will have more compact classes (fine grained classes) than others (classes with highly diverse examples). In practice, we compute three different reachability orderings for three different neighborhood sizes $V$ ($75, 50, 25$ neighbors), which allows us to account for cluster compactness variations across classes and noise levels. The algorithm chooses the optimal cluster assignment at the class level as being the cluster with the lower amount of outliers given at least two clusters are identified (clean and OOD). This allows us to reduce the amount of hyper-parameters to tune for the clustering to the $\xi$ parameter of OPTICS which controls the decision boundary between clusters and outliers. Higher values for $\xi$ imply a higher tolerance threshold meaning a lower amount of outliers.

\textbf{No ID noise.} In the case where we expect no ID noisy samples in the dataset, we only aim to discover a clean and an OOD cluster without outliers. In this case, the OOD cluster can be retrieved at the dataset level and we choose instead to fit a 2 component Gaussian mixture on the embedded features to retrieve the clusters.  

\textbf{Clean or OOD.} Once the clusters are evidenced in the ordering, the final step for the detection is to classify the clusters into clean or OOD. Although the average reachability score within the cluster could at first glance be considered a good indication of the OOD nature of a cluster, by computing the affinity matrix over a fixed neighborhood size, distances are not accurately preserved. We propose instead to compute the density of the cluster in the original unsupervised feature space, where for each sample in the dataset we compute the average distance to all other points in the cluster. We then select the cluster with the lowest density as the OOD cluster.

\subsection{Spectral Noise clustering from Contrastive Features (SNCF)}
Clustering the embedded unsupervised feature space provides three subsets of $\mathcal{X}$:  $\mathcal{X}_c, \mathcal{X}_n$ and $\mathcal{X}_o$, respectively the clean, ID noisy and OOD subsets. We aim to use all the available samples to train our CNN and do so by correcting ID noisy samples to their true label and using OOD samples to learn more robust low-level features. We train from scratch on each type of noise separately without using the unsupervised features to initialize the classification network.

\subsubsection{Correcting in-distribution noise}
Although the unsupervised features allow the detection of incorrectly assigned samples, we find that this is not sufficient to accurately assign ID noise to the right class, especially since they might be close to other ID noisy samples themselves assigned to another incorrect class. We propose instead to correct the ID noise during the supervised training phase, using knowledge learned on clean ID samples during a warm-up pre-training. We then estimate the true labels of the detected ID noise using a consistency regularization approach. For every ID noisy sample in $\mathcal{X}_n$ two weakly augmented versions are produced. The network then predicts on both samples ($p_{i,1}$ and $p_{i,2}$) and returns an average prediction, which, after temperature sharpening $\tau_1$ and normalization, is used as the corrected class assignment: $y_i = \left(\frac{p_{i,1} + p_{i,2}}{2}\right)^{\tau_1}$ with $\tau_1 = 2$ in every experiment. We find temperature sharpening to be necessary to reduce the entropy of the guessed label and to encourage the network to produce more confident predictions.

\subsubsection{Out-of-distribution samples}
OOD samples cannot be corrected to any label in the distribution but we propose to include them in an additional guided contrastive loss minimization objective to learn low level features. Once the noise detection algorithm has run, we re-embed the unsupervised features of detected OOD noise abd use OPTICS to discover clusters of the most similar samples in the OOD data. At training time, we augment each sample in the dataset into one weakly and one strongly augmented view, producing two mini-batches of the same images augmented differently. We then enforce ID samples belonging to the same class (corrected for the ID noise) as well as OOD samples from the same unsupervised cluster to be similar while being dissimilar to every other example in the mini-batch. OOD samples not assigned to any unsupervised cluster are considered similar to their augmented view only. This guided contrastive objective is described in equation~\ref{eq:guidedcont}. Note here that the similarities are enforced between the two mini-batches of augmented views alone. We attempted to enforce similarities inside the same augmented batch but noticed no accuracy improvements.

\begin{figure}[t]
\centering
\includegraphics[width=.95\linewidth]{"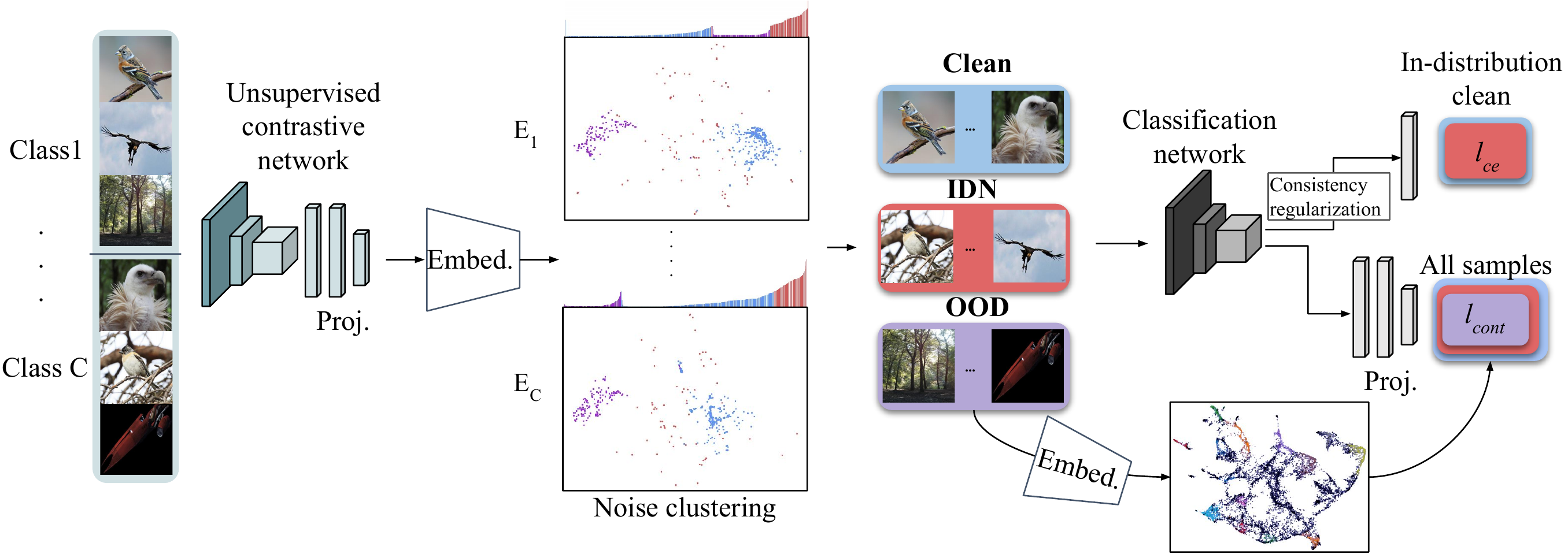"}
\par
\caption{Visualization of the algorithm. The unsupervised features are embedded to create $E$ and evaluated at the class level ($E_1,\dots E_C$) to cluster clean and OOD samples. The detected OOD samples are re-embedded from their unsupervised features to detect clusters of similar images. We correct the ID noise using consistency regularization and the OOD sample's cluster assignments are used together with the classes of all in-distribution samples in a guided contrastive objective \label{fig:algo}}
\end{figure}

\subsection{Loss objectives}
We consider here that $p_i$ is the current softmax prediction of $\Psi$ on sample $x_i \in \mathcal{X}$. Our algorithm aims to optimize over two objectives during training. The first is the classification objective on the detected clean samples $\mathcal{X}_c$ and the ID samples from $\mathcal{X}_n$ whose label has been corrected. We use the cross entropy loss:
\begin{equation}
    l_{ce} = \sum_{i = 1}^{N_c+N_n}y_i^T\log(p_i).~\label{eq:ID}
\end{equation}

Secondly, we minimize the guided contrastive learning objective, grouping samples of the same class and OOD samples from the same OOD cluster together using their respectively weakly and strongly augmented projected representations $r_i$ and $r_i'$, projected from the classification space to the contrastive space
\begin{equation}
    l_{cont} = - \frac{1}{N}\sum_{i=1}^{N}\frac{1}{B}\sum_{b=1}^{B}e_{i,b}\log\left(\frac{\exp{\left(\textit{ip}(r_b, r_i')/\tau_2\right)}}{\sum_{k=1}^B\exp{\left(\textit{ip}(r_k, r_i')/\tau_2\right)}}\right),~\label{eq:guidedcont}
\end{equation}
with $e_{i,b} = 1$ if sample $i$ is considered similar to sample $b$, $e_{i,b} = 0$ otherwise. Note that this objective can be paired with mixup as in the unsupervised objective in equation~\ref{eq:unsupcontmixup}.
The final loss minimized by our algorithm is:
\begin{equation}
    l = l_{ce} + \beta l_{cont},
\end{equation}
where $\beta$ is an hyper-parameter (typically $1$). Figure~\ref{fig:algo} illustrates the algorithm. 

\section{Experiments}
\subsection{Implementation details}
We form each mini-batch be aggregating an equal number of clean ID, noisy ID and OOD samples. Since the OOD is ignored in the ID objective (eq~\ref{eq:ID}) and in order to have the same batch size for the ID and the contrastive forward pass, we form the ID mini-batch by aggregating two weakly augmented views of the clean data with one weakly augmented view for the ID noisy data (ID clean, ID noisy, OOD for the contrastive mini-batch). For the weak data augmentations we use cropping with padding and random horizontal flip and the strong SimCLR augmentations~\cite{2020_ICML_SimCLR}. We warm-up the network on the detected clean data from scratch for $15$ epochs in every experiment (except $5$ for WebVision) and start both the ID noise correction and the guided contrastive objective after this. For a fair comparison with other approaches, the unsupervised features are not used to initialize the network in the robust classification phase. More experiments are proposed in the supplementary material. Since our algorithm minimizes a contrastive loss, we find that adding a non-linear projection head~\cite{2020_ICML_SimCLR,2020_arXiv_SimCLRv2} to project features from the classification space to the contrastive space is beneficial in reconciling the training objectives. The final number of projected contrastive features is $128$ and the projection head is not used at test time. We use stochastic gradient descent (SGD) with a weight decay of $5\times10^{-4}$ and mixup~\cite{2018_ICLR_mixup} augmentation with $\alpha = 1$ for all experiments. 

\textbf{Training the unsupervised algorithm.} We train the unsupervised algorithm using the same network as the robust classification phase. In cases where the resolution is $227\times 227$ or above, we train and evaluate the unsupervised features at resolution $84\times 84$ as this helps to keep training time and memory consumption reasonable yet still separates the OOD and ID clusters. The algorithm is trained for $2000$ epochs, with a batch size of $256$, starting with a learning rate of $0.01$ and reducing it by a factor of $10$ at epochs $1000, 1500$. We use the mixup version on the unsupervised objective (iMix~\cite{2021_ICLR_iMix}).

\textbf{Synthetically corrupted datasets.} We conduct a first series of experiments on synthetically corrupted versions of CIFAR-100~\cite{2009_CIFAR} where we control the ID noise and OOD noise. We use the same configuration as in Albert~\etal~\cite{2022_WACV_DSOS} and note $r_{in}$ and $r_{out}$ the corruption ratios for ID noisy and OOD noise respectively with $r_{in} + r_{out}$ the total noise level. Our focus here is on the OOD noise rate more than ID noise, which is less present in web-crawled datasets~\cite{2022_WACV_DSOS}. We introduce OOD noise by replacing original images with images from another dataset, either ImageNet32~\cite{2017_arXiv_IN32} or Places365~\cite{2017_TPAMI_places}. For the ID noise, we randomly flip the labels of a portion of the dataset to a random label (uniform noise). The dataset size remains 50K images after noise injection. We train on CIFAR-100 using a PreActResNet18~\cite{2016_CVPR_ResNet} trained with a batch size of $256$ for $100$ epochs with a learning rate of $0.1$, reducing it by a factor of $10$ at epochs $50$ and $80$. 

\textbf{Web noise corruption.} We conduct experiments on miniImageNet~\cite{2016_NIPS_MiniImageNet} corrupted by web noise from Jiang~\etal~\cite{2020_ICML_MentorMix} (Controlled Noisy Web Labels, CNWL) where the severity of the web noise corruption is controlled. This dataset is an example where ID noise is very limited and where we find that using the 2 components GMM is sufficient to detect the noise at the dataset level (see Section~\ref{sec:unsupclust}). We train on this dataset at two different resolutions, first $299\times 299$, which is the original configuration proposed by Jiang~\etal~\cite{2020_ICML_MentorMix} and second the $32\times32$ resolution adopted in recent works~\cite{2021_arXiv_propmix,2021_CVPR_FaMUS,2021_arXiv_Scanmix}. For the $299\times299$ configuration, we train an InceptionResNetV2~\cite{2016_AAAI_Inception} with a batch size of $64$ for $200$ epochs with a learning rate of $0.01$, reducing it by a factor of $10$ at epochs $100, 160$. For the $32\times32$ configuration, we use the same configuration as CIFAR-100.

\textbf{Real-world dataset.} We evaluate our model on the (mini)Webvision~\cite{2017_arXiv_WebVision} dataset reduced to the first 50 classes (65k images). We train an InceptionResNetV2~\cite{2016_AAAI_Inception} at a $227\times227$ resolution with a batch size of $64$ for $100$ epochs with a learning rate of $0.01$, reducing it by $10$ at epochs $50, 80$.

\textbf{Baselines.~\label{par:baselines}} We introduce here the state-of-the-art approaches we compared with as well as the abbreviations used in the tables. Cross-entropy (CE), dropout (D), and mixup (M) are simple baselines obtained by training with no noise correction and dropout~\cite{2017_ICML_dropout} or mixup~\cite{2018_ICLR_mixup} as regularization. MentorNet~\cite{2018_ICML_MentorNet} (MN) and MentorMix~\cite{2020_ICML_MentorMix} (MM) use teacher networks to weight noisy samples. FaMUS~\cite{2021_CVPR_FaMUS} (FaMUS) uses meta learning to learn to correct noisy samples.  Bootstrapping~\cite{2015_ICLR_Bootstrapping} (B) corrects noisy samples using a fixed interpolation with pseudo-labels; Dynamic Bootstrapping~\cite{2019_ICML_BynamicBootstrapping} (DB) expands the idea by correcting only high loss noisy samples retreived using a beta mixture. The S-model~\cite{2017_ICLR_Smodel} (SM) corrects noisy samples using a noise adaptation layer optimized using an expectation maximization algorithm. DivideMix~\cite{2020_ICLR_DivideMix} (DM) uses a Gaussian mixture to detect high loss samples and correct them using a semi-supervised consistency regularization algorithm; the idea is expanded upon in PropMix~\cite{2021_arXiv_propmix} (PM) where self-supervised initialization is used and only the simplest of the noisy samples are corrected while the hardest are discarded. ScanMix~\cite{2021_arXiv_Scanmix} (SM) also improves on DM by correcting the label using a semi-supervised contrastive algorithm together with a semantic clustering in an self-supervised feature space, optimized using an EM algorithm. EvidentialMix~\cite{2020_WACV_EDM} (EDM) refines the noisy sample detection of DM to account for OOD samples and uses the evidential loss~\cite{2018_NeurIPS_evidentialloss} to evidence separate OOD and ID noisy modes. JoSRC~\cite{2021_CVPR_JoSRC} (JoSRC) proposes to use the Jensen-Shannon divergence between a consistency regularization guessed label and the original label to detect noisy samples and further select samples with low agreement between views as OOD. Robust Representation Learning~\cite{2021_ICCV_RRL} (RRL) trains a weakly supervised prototype objective to promote clean samples to be close to their class prototypes. Finally, Dynamic Softening for Out-of-distribution Samples~\cite{2022_WACV_DSOS} (DSOS) computes the collision entropy of the interpolation between the original label and network prediction to separate ID noisy and OOD samples.
\begin{figure}[t]
\centering
\includegraphics[width=.6\linewidth]{"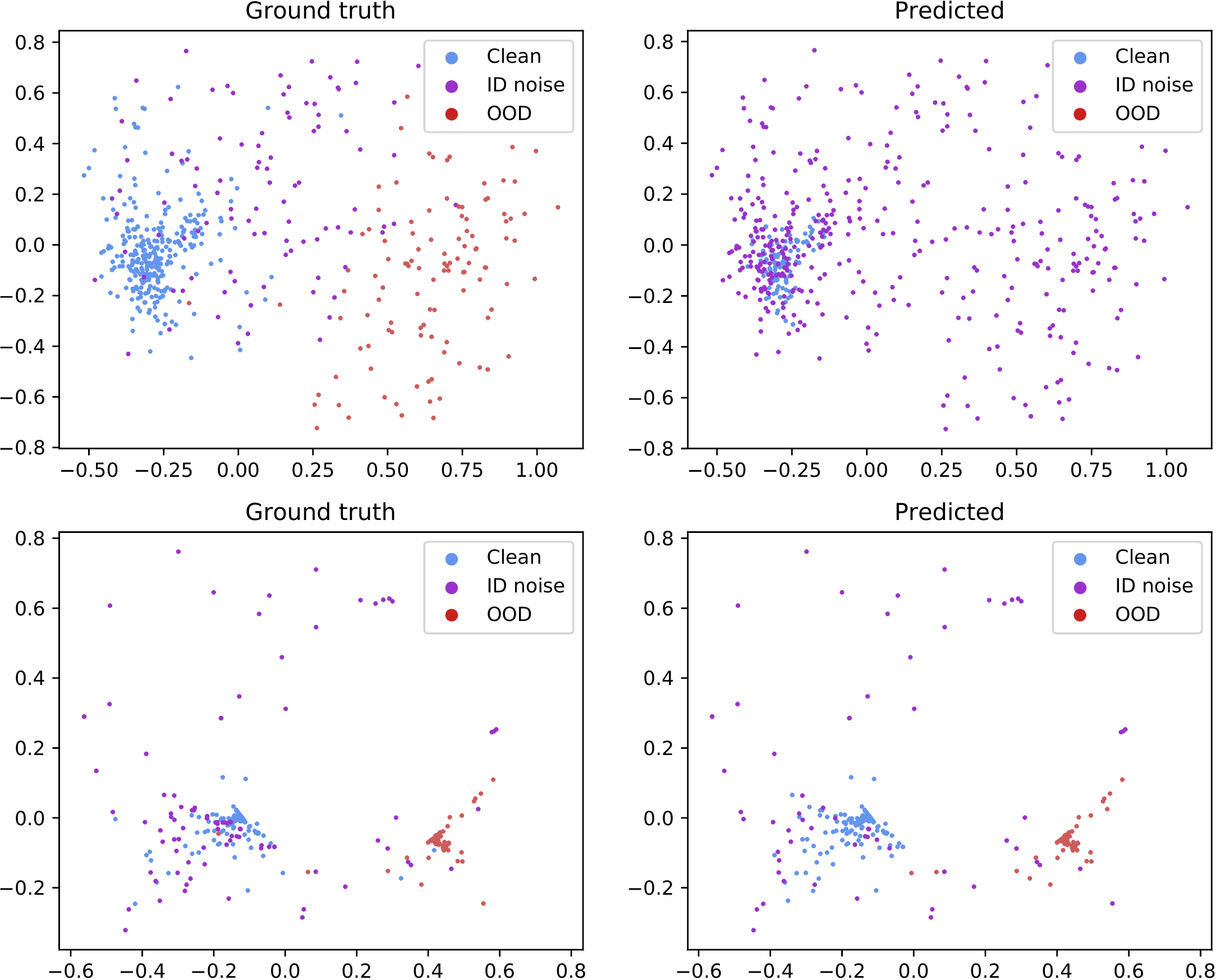"}
\par
\caption{Feature embedding for class $1$ of CIFAR-100 corrupted with $r_i = r_o = 0.2$ (ImageNet32 OOD). The top row presents a 2D visualization obtained using Isomap~\cite{2000_Science_Isomap} of the raw contrastive features and the second row presents a Isomap visualization of the embedding E. Embedding the features allows to evidence the OOD cluster \label{fig:visembed}}
\end{figure}

\subsection{Clustering the unsupervised features~\label{sec:linearsep}}
This section presents the experiments on the linear separability of ID and OOD data in an unsupervised feature space, the importance of embedding the unsupervised features when clustering the noise and the accuracy our noise retrieval algorithm. First, to validate our hypothesis over the separability of ID and OOD samples in the unsupervised feature space, we propose to train a linear classifier on the unsupervised features and evaluate its capacity to linearly separate the two distributions. We observe that the classifier can linearly separate the two distribution with error rates below $3\%$ for synthetically corrupted CIFAR-100 with $r_{in}=r_{out}=0.2$ and below $1\%$ for miniImagenet corrupted with web noise (CNWL). The linear separability is less accurate when using Places365 as the OOD corruption dataset; we argue that this is because of lower image variability in the dataset, justified by the lower number of classes and the fine-grained nature of the classification task. A table and more visualizations of the separation on the 2D-hypersphere are available in the supplementary material.
Second, Figure~\ref{fig:visembed} provides a visualization of the importance of embedding the unsupervised features to perform the noise clustering for a class of CIFAR-100 where we compare applying the clustering algorithm on the raw unsupervised contrastive features against the spectral embedding $E$. The left column is the ground-truth and the right represent the detection made by the clustering algorithm. We use Isomap~\cite{2000_Science_Isomap} to reduce the dimentionality to $2$ to be able to visualize the features. The spectral embedding $E$ is essential to evidence the OOD cluster, not originally present in the raw features.

\subsection{Synthetic noise corruption}
We study the capacity of our algorithm to mitigate ID noise and OOD noise on synthetically corrupted version of the CIFAR-100 dataset. Table~\ref{tab:sotac100} reports results when using ImageNet32 or Places365 as a OOD corruption. We notice that the OOD corruption using the Places365 dataset is more harmful than corrupting with ImageNet32, especially for high noise levels.

\setlength{\tabcolsep}{3pt}
\begin{table}[t]
    \caption{Mitigating ID noise and OOD noise on CIFAR-100 corrupted with ImageNet32 or Places365 images. We run all the algorithms using publicly available implementations by authors. We report best and last accuracy. We bold (underline) the highest best (final) accuracy
    \label{tab:sotac100}}
    \global\long\def\arraystretch{0.9}%
    \centering
    \resizebox{1\textwidth}{!}{%
    \centering
    \begin{tabular}{c>{\centering}c>{\centering}c>{\centering}c>{\centering}c>{\centering}c>{\centering}c>{\centering}c>{\centering}c>{\centering}c>{\centering}c>{\centering}c}
    \toprule
    Corruption & $r_{out}$ & $r_{in}$ & CE & M & DB & JoSRC & ELR & EDM & {DSOS} & RRL & {Ours} \tabularnewline
    \midrule
    \multirow{4}{*}{INet32}& $0.2$ & $0.2$ & $63.68/55.52$ & $66.71/62.52$ & $65.61/65.61$ & $67.37/64.17$ & $68.71/68.51$ & $71.03/70.42$ & $70.54/70.54$ & $72.64/72.33$ & $\textbf{72.95}/\underline{72.70}$ \tabularnewline
    & $0.4$ & $0.2$ & $58.94/44.31$ & $59.54/53.16$ & $54.79/54.42$ & $61.70/61.37$ & $63.21/63.07$ & $61.89/61.83$ & $62.49/62.05$ & $66.04/65.44$ & $\textbf{67.62}/\underline{67.14}$ \tabularnewline
    & $0.6$ & $0.2$ & $46.02/26.03$ & $42.87/40.39$ & $42.50/42.50$ & $37.95/37.11$ & $44.79/44.60$ & $21.88/14.59$ & $49.98/49.14$ & $26.76/24.51$ & $\textbf{53.26}/\underline{51.26}$  \tabularnewline
    & $0.4$ & $0.4$ & $41.39/18.45$ & $38.37/33.85$ & $35.90/35.90$ & $41.53/41.44$ & $34.82/34.21$ & $24.15/01.62$ & $43.69/42.88$ & $31.29/30.64$ & $\textbf{54.04}/\underline{52.66}$ \tabularnewline
    \midrule
    \multirow{4}{*}{Places365}& $0.2$ & $0.2$ & $59.88/53.61$ & $66.31/59.69$ & $65.86/65.83$ & $67.06/66.73$ & $68.58/68.45$ & $70.46/70.25$ & $69.72/69.12$ & $\textbf{72.62}/\underline{72.49}$ & $71.25/71.14$ \tabularnewline
    & $0.4$ & $0.2$ & $53.46/42.46$ & $59.75/48.55$ & $55.81/55.61$ & $60.83/60.64$ & $62.66/62.34$ & $61.80/61.55$ & $59.47/59.47 $ & $\textbf{65.82}/\underline{65.79}$ & $64.03/63.48$\tabularnewline
    & $0.6$ & $0.2$ & $39.55/21.42$ & $ 39.17/33.69$ & $40.75/40.61$ & $39.83/39.63$ & $37.10/36.51$ & $23.67/14.66$ & $35.48/35.41$ &  $49.27/49.27$ & $\textbf{49.83}/\underline{49.83}$\tabularnewline
    & $0.4$ & $0.4$ & $32.06/13.85$ & $34.36/27.63$ & $35.05/34.86$ & $33.23/32.58$ & $34.71/33.86$ & $20.33/11.88$ & $29.54/29.48$ & $26.67/24.34$ & $\textbf{50.95}/\underline{47.61}$\tabularnewline
       \bottomrule
    \end{tabular}}
    
\end{table}

\begin{table}[t]
    \caption{Ablation study on CIFAR-100 corrupted with ImageNet32 with $r_{out} = 0.4$ and $r_{in} = 0.2$. corr = correction and rm = remove \label{tab:abla}}
    \global\long\def\arraystretch{1}%
    \centering
    \resizebox{.6\textwidth}{!}{{{}}%
    \begin{tabular}{l l>{\centering}c>{\centering}c>{\centering}c>{\centering}c}
    \toprule
        & & Embed & Contrastive  & Best & Last\tabularnewline
        \midrule
       \multirow{3}{*}{No noise corr} & CE & \xmark & \xmark & $58.94$ & $44.31$ \tabularnewline
       & + mixup & \xmark & \xmark &$59.54$ & $53.16$ \tabularnewline
       & + guided contrastive & \xmark & \cmark & $62.83$ & $56.29$ \tabularnewline
       \midrule
       \multirow{10}{*}{Noise corr} & ID corr only &  \xmark & \xmark &  57.02 & 55.43 \tabularnewline
       & rm OOD only &  \xmark & \xmark & 60.73 & 53.88 \tabularnewline
       & ID corr and rm OOD &   \xmark & \xmark & 54.81 & 54.20 \tabularnewline
       \cmidrule(lr){2-6}
       & ID corr only &  \cmark & \xmark & 61.40 & 58.90 \tabularnewline
       & rm OOD only &  \cmark & \xmark & 60.87 & 54.08 \tabularnewline
       & ID corr and rm OOD &   \cmark & \xmark & 61.83 & 61.45 \tabularnewline
       \cmidrule(lr){2-6}
         & ID corr only &  \cmark & \cmark & 63.91 & 62.94 \tabularnewline
       & ID corr and rm OOD &   \cmark & \cmark & 64.51 & 64.04 \tabularnewline
       & OOD corr only &  \cmark & \cmark & 63.41 & 58.39 \tabularnewline
       & ID + OOD corr & \cmark & \cmark & 65.22 & 64.42 \tabularnewline
       \midrule
       \multirow{2}{*}{Other}  &+ equal sampling & \cmark & \cmark & 67.62 & 67.14 \tabularnewline
       & - mixup & \cmark & \cmark & 61.66 & 59.40  \tabularnewline
       \bottomrule
    \end{tabular}}

\end{table}

\subsection{Ablation study}
\begin{figure}[t]
\centering
\includegraphics[width=.8\linewidth]{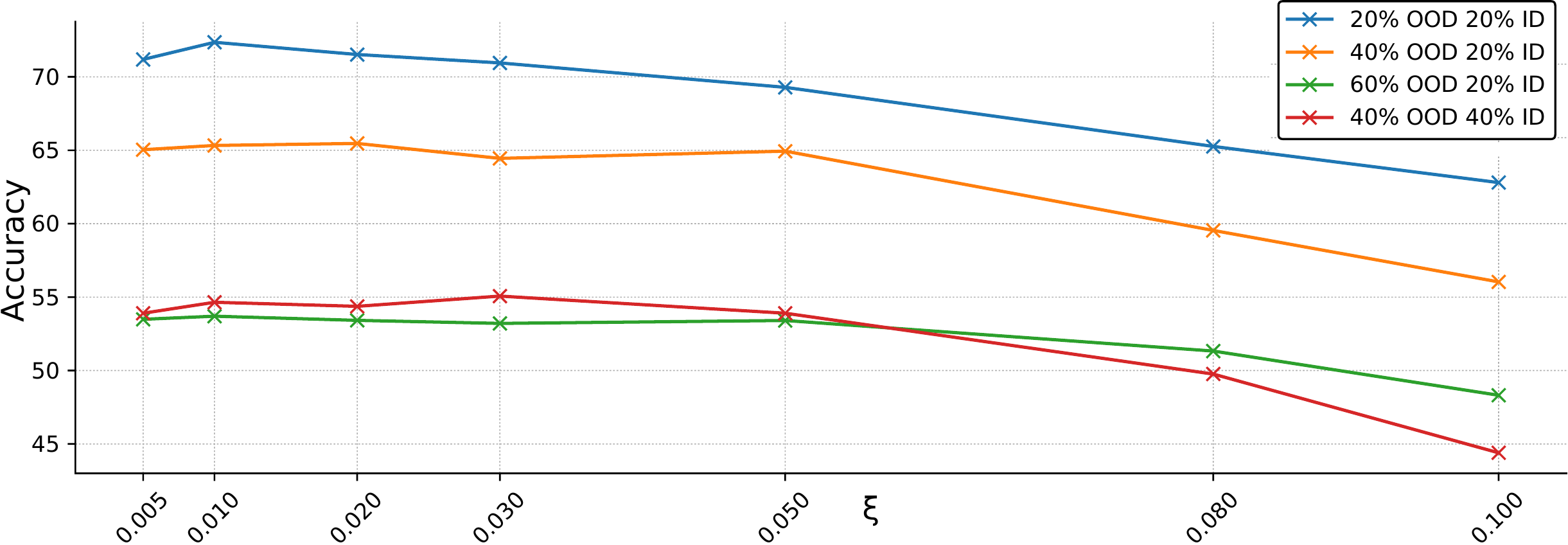}
\par
\caption{Hyper-parameter tuning for OPTICS. We report accuracy results obtained for ID/OOD clustering setting different $\xi$ values in OPTICS\label{fig:xi}}
\end{figure}

Table~\ref{tab:abla} illustrates the importance of each element of the proposed method on CIFAR-100 corrupted with OOD noise from ImageNet32. We study multiple cases including retrieving OOD and ID clusters on the un-embedded raw unsupervised contrastive features (Noise corr without embedding); correcting only the OOD or ID examples while considering the rest clean (ID/OOD corr only); joint effect of the ID and OOD correction (ID + OOD corr); studying the effect of removing the OOD samples from the training set instead of using them in the guided contrastive objective in equation~\ref{eq:guidedcont} (ID corr and rm OOD); runing the algorithm without mixup (- mixup). We point out how important mixup is (especially in the classification loss) to avoid overfitting to the noise. Figure~\ref{fig:xi} reports the quality of our robust classification algorithm on ID and OOD clustering for different values of $\xi$ in OPTICS where values inferior to $0.03$ lead to the best results. We choose $\xi=0.01$ for all datasets.

\begin{table}[t]
    \caption{Web-corrupted miniImageNet from the CNWL~\cite{2020_ICML_MentorMix} ($32\times 32$). We run our algorithm; other results are from~\cite{2021_arXiv_propmix}. We denote with $\star$ algorithms using an ensemble of networks to predict and with $\dagger$ algorithms using unsupervised initialization. We report best accuracy
    \label{tab:sotamini32}}
    \global\long\def\arraystretch{0.9}%
    \centering
    \resizebox{.6\textwidth}{!}{%
    \centering
    \begin{tabular}{c>{\centering}c>{\centering}c>{\centering}c>{\centering}c>{\centering}c>{\centering}c>{\centering}c>{\centering}c>{\centering}c>{\centering}c}
    \toprule
    Noise level & CE & M & $\star$DM & MM & FaMUS & $\star\dagger$SM & $\star\dagger$PM & Ours \tabularnewline
    \midrule
    20 & $47.36$ & $49.10$ & $50.96$ & $51.02$ & $51.42$ &  $59.06$ & $61.24$ & $61.56$ \tabularnewline
    40 & $42.70$ & $46.40$ & $46.72$ & $47.14$ & $48.03$ & $54.54$ & $56.22$ & $59.94$ \tabularnewline
    60 & $37.30$ & $40.58$ & $43.14$ & $43.80$ & $45.10$ &
    $52.36$ &  $52.84$ & $54.92$  \tabularnewline
    80 & $29.76$ & $33.58$ & $34.50$ & $33.46$ & $35.50$ & $40.00$ & $43.42$ & $45.62$ \tabularnewline
       \bottomrule
    \end{tabular}}
     
\end{table}

\begin{table}[t]
    \caption{Web noise on web corrupted miniImageNet (Red MiniImageNet~\cite{2020_ICML_MentorMix}) trained at a high resolution ($299\times 299$). We run our algorithm, other results are from~\cite{2022_WACV_DSOS}. We report best accuracy
    \label{tab:sotamini}}
    \centering
    \global\long\def\arraystretch{0.9}%
    \resizebox{.9\textwidth}{!}{{{}}%
    \begin{tabular}{l>{\centering}c>{\centering}c>{\centering}c>{\centering}c>{\centering}c>{\centering}c>{\centering}c>{\centering}c>{\centering}c>{\centering}c}
    \toprule
    Noise level & CE & D & SM & B & M & MN & MM & DSOS & Ours\tabularnewline
    \midrule
    $0$ & $70.9/68.5$ & $71.8/65.7$ & $71.4/68.4$ & $71.8/68.4$ & $72.8/72.3$ & $71.2/68.9$ & $74.3/73.7$ & $74.52/74.10$ & $\textbf{74.80}/\underline{74.60}$\tabularnewline
    $30$ & $66.1/56.5$ & $66.6/55.0$ & $65.2/56.3$ & $66.6/56.7   $ &$66.8/61.8$ & $66.2/64.0$ & $68.3/67.2$ & $69.84/67.86$ & $\textbf{69.96}/\underline{69.64}$\tabularnewline
    $50$ & $60.9/51.7$ & $62.1/50.01$ & $61.3/51.3$ & $62.6/52..5$ & $63.2/58.4$ & $61.7/58.0$ & $63.3/61.8$ & $66.14/65.18$ & $\textbf{66.48}/\underline{66.38} $\tabularnewline
    $80$ & $48.8/39.8$ & $49.5/37.6$ & $49.0/40.6$ & $50.1/40.1$ & $50.7/45.5$ & $49.3/43.4$ & $50.2/48.4$ & $55.26/52.24$ & $\textbf{55.54}/\underline{54.96} $\tabularnewline
    \bottomrule
    \end{tabular}}
\end{table}

\begin{table}[t]
    \caption{Classification accuracy for the proposed and other state-of-the-art methods. We denote with $\star$ algorithms using an ensemble of networks to predict and with $\dagger$ algorithms using unsupervised initialization.. We train the network on the mini-Webvision dataset and test on the ImageNet 1k test set (ILSVRC12).  We bold the best results \label{tab:webvis}}
    \centering
    \global\long\def\arraystretch{0.9}%
    \resizebox{.9\textwidth}{!}{{{}}%
    \begin{tabular}{l>{\centering}c>{\centering}c>{\centering}c>{\centering}c>{\centering}c>{\centering}c>{\centering}c>{\centering}c>{\centering}c>{\centering}c>{\centering}c>{\centering}c>{\centering}c>{\centering}c}
    \toprule
        & & \multicolumn{10}{c}{100 epochs} & \multicolumn{2}{c}{150 epochs}  \tabularnewline
        \cmidrule(lr){3-12} \cmidrule(lr){13-14}
        & & M & MM & $\star$DM & $\star$ELR+ & RRL & $\star$DSOS & $\star\dagger$PM  &  $\star\dagger$SM & Ours & $\star$Ours & FaMUS & $\star$Ours \tabularnewline
    \midrule
       \multirow{2}{*}{mini-WebVision} & top-1 & $75.44$ & $76.0$& $77.32$ & $77.78$ & $77.80$ & $78.76$ & $78.84$ & $\mathbf{80.04}$ & $78.16$ & $79.84$ & $79.40$ & $\mathbf{80.24}$\tabularnewline
        & top-5 & $90.12$ & $90.2$ & $91.64$ & $91.68$ & $91.30$ & $92.32$ & $90.56$  & $93.04$ & $92.60$ & $\mathbf{93.64}$ & $92.80$ & $\mathbf{93.44}$\tabularnewline
       \multirow{2}{*}{ILSVRC12} & top-1 & $71.44$ & $72.9$& $75.20$ & $70.29$ & $74.40$ & $75.88$ & $--$ & $75.76$ & $74.20$ & $\mathbf{76.64}$ & $77.00$ & $\mathbf{77.12}$\tabularnewline
       & top-5 & $89.40$ & $91.10$ & $90.84$ & $89.76$ & $90.90$ & $92.36$& $--$&  $92.60$ & $93.32$ & $\mathbf{94.20}$ & $92.76$ & $\mathbf{94.32}$\tabularnewline
       \bottomrule
    \end{tabular}}
     
\end{table}

\subsection{Results on web-noise}
We consider here the controlled noisy web labels (CNWL) dataset, where miniImageNet is corrupted with OOD images from web queries. Table~\ref{tab:sotamini32} reports results when training at resolution $32\times32$ and Table~\ref{tab:sotamini} at resolution $299\times299$. Finally, in Table~\ref{tab:webvis} we train on the first 50 classes of the Webvision dataset (mini-Webvision) a real world web-crawled dataset and report top-1 and top-5 accuracy results on the validation set on Webvision and on the test set on the ImageNet1k (ILSVRC12) dataset. Since our algorithm uses only one network and to compare against ensemble methods, we report an additional result where we ensemble two networks trained from different random initializations. We also report results when training for $150$ epochs to compare fairly against FaMUS. Our algorithm slightly outperforms the state-of-the-art for top-1 accuracy but more convincingly so for top-5 accuracy on Webvision. Because of the guided contrastive loss, the network learns more generalizable features which reduce the risk of catastrophic classification errors (when the predicted class is completely semantically different from the correct predictions). mini-Webvision in particular proposes fine grained classification on species of birds, amphibians and marine animals which reward generalizable features for top-5 evaluation. 

\section{Conclusion}
This paper proposes to use the alignment and uniformity principles of unsupervised contrastive learning to detect clean and OOD label noise clusters in an embedded feature space. We show that the unsupervised contrastive features for OOD and ID samples are, to a large extent, linearly separated on the unit hypersphere and compute a fixed neighborhood spectral embedding to reduce differences in cluster densities. We adapt the OPTICS algorithm, ordering samples in a neighbor chain and computing the reachability cost to neighbors. Clusters are evidenced by valleys in the reachability plot and a voting system automatically selects the best cluster assignement at the class level given multiple neighborhood sizes. Once the noise has been identified, we train a robust classifier that corrects the labels of known ID noisy samples using a consistency regularization estimation and uses ID and OOD samples together in an auxiliary guided contrastive objective. We report state-of-the-art results on a variety of noisy datasets including synthetically corrupted versions of CIFAR-100, controlled web noise in miniImageNet, and mini-Webvision as a real-world web-crawled dataset.
\iftrue
\section*{Acknowledgments} This publication has emanated from research conducted with the financial support of Science Foundation Ireland (SFI) under grant number SFI/15/SIRG/3283 and SFI/12/RC/2289\_P2 as well as the support of the Irish Centre for High End Computing (ICHEC).
\fi

\clearpage
%
%
\bibliographystyle{splncs04}
\bibliography{egbib}
\end{document}


\pagestyle{headings}
\mainmatter
\def\ECCVSubNumber{6827}  
\title{Supplementary material: Embedding contrastive unsupervised features to cluster in- and out-of-distribution noise in corrupted image datasets}

\titlerunning{Supplementary material: Embedding features to cluster OOD noise}
\author{
\institute{}
}
\authorrunning{Albert~\etal} 
\maketitle

\section{Linear separability of samples on the hyper-sphere}
\begin{figure}
\centering
\includegraphics[width=.99\linewidth]{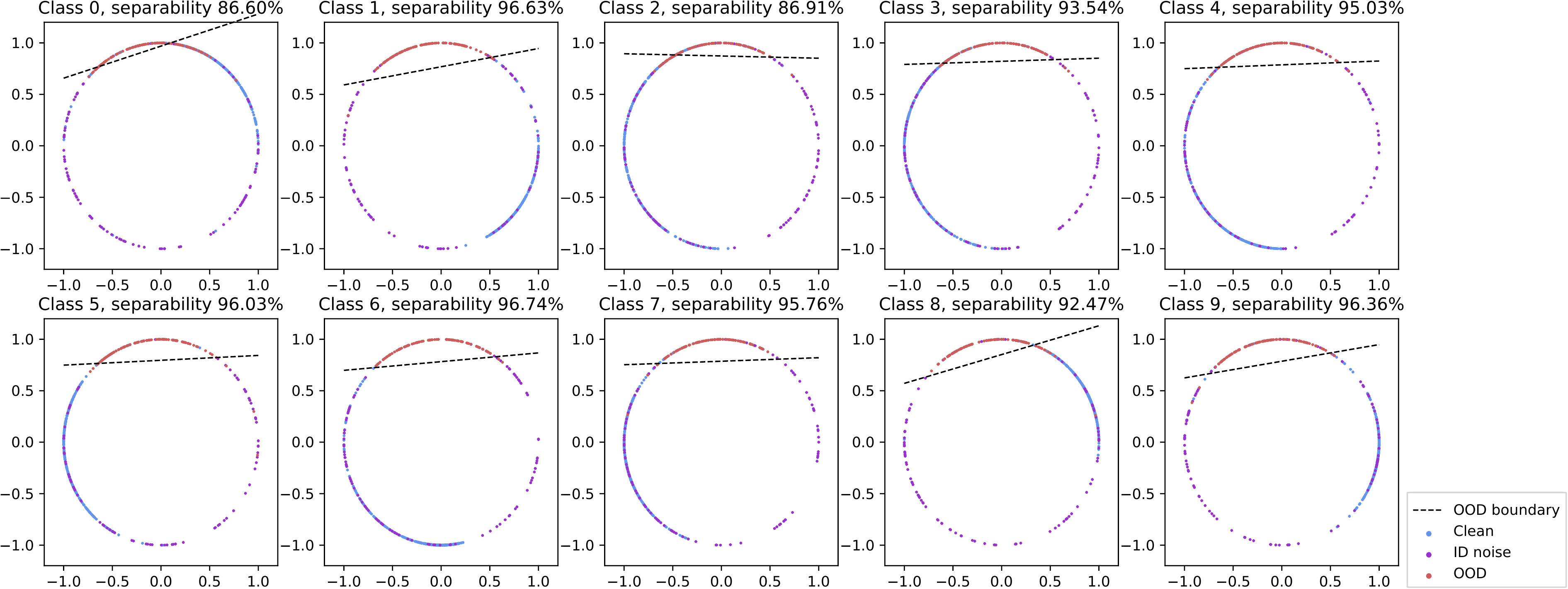}
\par
\caption{Linear separation of OOD and ID on the hypersphere for each of the CIFAR-10 classes corrupted with $r_{in} = r_{out} = 0.2$, OOD from ImageNet32. Linear separability at the class level. \label{fig:cifar10class}}
\end{figure}

Figure~\ref{fig:cifar10class} illustrates the linear separability of samples on the hypersphere on for each class of CIFAR-10 corrupted with ID noise and OOD noise from ImageNet32. We train the unsupervised $N$-pairs algorithm and use a non-linear projection with the final dimension being $2$ as in Wang~\etal~\cite{2020_ICLR_understandingcontrastive}. Here we use the simpler CIFAR-10 dataset because the final 2D projection size is too small, which causes convergence issues with more difficult classification problems such as CIFAR-100. The linear separation is not as good as when using the larger dimension of $128$ for the contrastive projection head but allows direct visualization of the separation. We display $1,000$ randomly selected samples at the dataset level as well as the predicted linear boundary. The OOD samples cluster on one side of the circle, confirming our hypothesis by becoming linearly separable from the in-distribution data. 

\setlength{\tabcolsep}{3pt}

\section{Unsupervised initialization for label noise robust algorithms}
In this paper, we chose to use the unsupervised features to detect the label noise but to avoid using them to initialize the CNN in the supervised phase. We did this to provide a fair comparison with existing noise robust algorithms as we observe that naive unsupervised initialization will harm the detection of out-of-distribution noise. Although existing works have showed that unsupervised learning can be used to improve robustness to in-distribution noise~\cite{2021_arXiv_propmix,2019_ICCV_SelfLearningLabelNoise} or to reduce uncertainty~\cite{2019_NeurIPS_SelfsupRobustness} we find that the effects are detrimental in the presence of OOD noise. We believe that this is because unsupervised learning will learn features for OOD samples before the supervised phase, making it easier to overfit the OOD noise and reduce the capacity of existing algorithms to detect OOD images since all existing approaches rely on CNNs producing underconfident predictions on OOD samples~\cite{2021_ICCV_RRL,2021_CVPR_JoSRC}. Table~\ref{tab:unsupinit} reports accuracy results of three state-of-the-art algorithms (EDM~\cite{2020_WACV_EDM}, DSOS~\cite{2022_WACV_DSOS}, RRL~\cite{2021_ICCV_RRL}) and our algorithm trained using a CNN trained from random initialization or initialized using iMix~\cite{2021_ICLR_iMix} when trained on CIFAR-100 corrupted with $20\%$ of ID and OOD noise from ImageNet32. We also report baselines that do not perform noise or label correction: CE and M~\cite{2018_ICLR_mixup}. We find that noise robust algorithms designed to function from a random initialization perform much worse when a self-supervised initialization is used. Note that part of the accuracy decrease could also be due to hyper-parameters that we did not tune.  Standard cross-entropy training (CE) and Mixup (M) perform slightly worse as well or at least does not benefit from the unsupervised weight initialization. We encourage future research in this direction.
 
\begin{table}[t]
    \caption{Self-supervised initialization harms OOD robustness. CIFAR-100 corrupted with $r_{in} = r_{out} = 0.2$
    \label{tab:unsupinit}}
    \global\long\def\arraystretch{0.9}%
    \centering
    \resizebox{.7\textwidth}{!}{%
    \centering
    \begin{tabular}{c>{\centering}c>{\centering}c>{\centering}c>{\centering}c>{\centering}c>{\centering}c>{\centering}c>{\centering}c>{\centering}c}
    \toprule
    Noise &  Unsup init & CE & M & EDM & DSOS & RRL & Ours \tabularnewline
    \midrule
    $r_{in} = 0.2$ &  \xmark & $65.79$ & $67.50$ & $71.03$ & $70.54$ & $72.64$ & $72.95$ \tabularnewline
    $r_{out} = 0.2$ & \cmark & $64.89$ & $66.99$ & $56.89$ & $66.76$ & $66.76$ & $71.62$ \tabularnewline
    \bottomrule
    \end{tabular}}
    
\end{table}

\begin{table}[t]
    \caption{Linear separability between ID and OOD samples on the hypersphere in CIFAR-100 using ImageNet32 or Places365 as OOD data and on the miniImageNet part of CNWL dataset.
    \label{tab:linearsep}}
    \global\long\def\arraystretch{0.9}%
    \centering
    \resizebox{.7\textwidth}{!}{%
    \centering
    \begin{tabular}{c>{\centering}c>{\centering}c>{\centering}c>{\centering}c>{\centering}c>{\centering}c>{\centering}c>{\centering}c>{\centering}c}
    \toprule
    Dataset & \multicolumn{2}{c}{Cifar-100} & \multicolumn{2}{c}{miniImageNet} \tabularnewline
    \midrule
    Corruption dataset & INet32 & P365 & Web & Web \tabularnewline
    $r_{out}$ & $0.2$ & $0.2$ & $0.2$ & $0.6$ \tabularnewline
    \midrule
    Linear classifier score & $98.21$ & $95.95$ & $99.66$ & $99.54$ \tabularnewline
       \bottomrule
    \end{tabular}}
    
\end{table}
\section{Hyper-parameter table for experiments}
\begin{table*}
    \caption{hyper-parameters for training SNCF on corrupted image datasets. \label{tab:hyper}}
    \global\long\def\arraystretch{0.9}%
    \centering
    \resizebox{1\textwidth}{!}{%
    \centering
    \begin{tabular}{l>{\centering}c>{\centering}c>{\centering}c>{\centering}c>{\centering}c>{\centering}c>{\centering}c>{\centering}c>{\centering}c>{\centering}c}
    \toprule
    Dataset & corruption & resolution & $\beta$ & epochs & batch size & lr & lr red. & warmup & mixup & network \tabularnewline
    \midrule
    \multirow{2}{*}{CIFAR-100} & INet & $32\times32$ & $1$ & $100$ & $256$ & $0.1$ & $50, 80$ & $15$ & yes & PreActRes18 \tabularnewline
     & Places & $32\times32$ & $1$ & $100$ & $256$ & $0.1$ & $50, 80$ & $30$ & yes & PreActRes18 \tabularnewline
     \midrule
    \multirow{2}{*}{CNWL} & \multirow{2}{*}{Web} & $32\times32$ & $\tilde{r}_{out}$ & $100$ & $256$ & $0.1$ & $50, 80$ & $15$ & yes & PreActRes18 \tabularnewline
    & & $299\times299$ & $1$ & $200$ & $64$ & $0.01$ & $100, 160$ & $15$ & yes & InceptionResNetV2 \tabularnewline
    \midrule
    Webvision & Web & $227\times227$ & $1$ & $100$ & $64$ & $0.01$ & $50, 80$ & $5$ & yes & InceptionResNetV2 \tabularnewline
    \bottomrule
    \end{tabular}}
\end{table*}
Table~\ref{tab:hyper} references the different hyper-parameters we use to train SNCF. We find that in most cases $\beta=1$ is a sufficient hyper-parameter except for the CNWL dataset in the $32\times32$ resolution where we use the estimated ratio of OOD samples in the dataset $\tilde{r}_{out}$. This is not necessary for the $299\times299$ resolution where we use $\beta=1$ for all noise configurations.

\section{Run times}
Although OPTICS is computationally expensive, we only run it once at the beginning of training on the embedded unsupervised features so it has no impact on the epoch train time. For reference, computation time for the spectral embedding + three OPTICS iterations for different neighborhood sizes \{$25, 50, 75$\} on an i7-8700K in Python (averaged over 10 runs) takes 93s for 50,000 samples (100 classes) and 1h19m for 1,000,000 samples (1000 classes). This one-off cost is offset because we skip extra forward passes computed every epoch by other label robust algorithms (EDM, DSOS, SM) to evaluate feature representations or losses throughout  training. Single epoch training times on a RTX 2080TI on CIFAR-100 with 20\% OOD and ID noise are: ours 63s, EDM 93s, and DSOS 57s. Because of the equal sampling we perform in the algorithm, lower noise levels lead to longer run times as noisy examples are over-sampled to complete clean batches. This low noise configuration is the longest run time for SNCF. All algorithms are run using half precision. 

\section{Hardware}
All networks are trained using mixed precision on a Nvidia RTX 2080TI ($32\times32$ and $84\times84$ resolution) and two Nvidia TeslaV100 ($227\times 227$ and $299\times 299$ resolution)
\section{Visualization of OOD sample clustering on the CNWL}
\begin{figure*}
\centering
\includegraphics[width=.9\linewidth]{"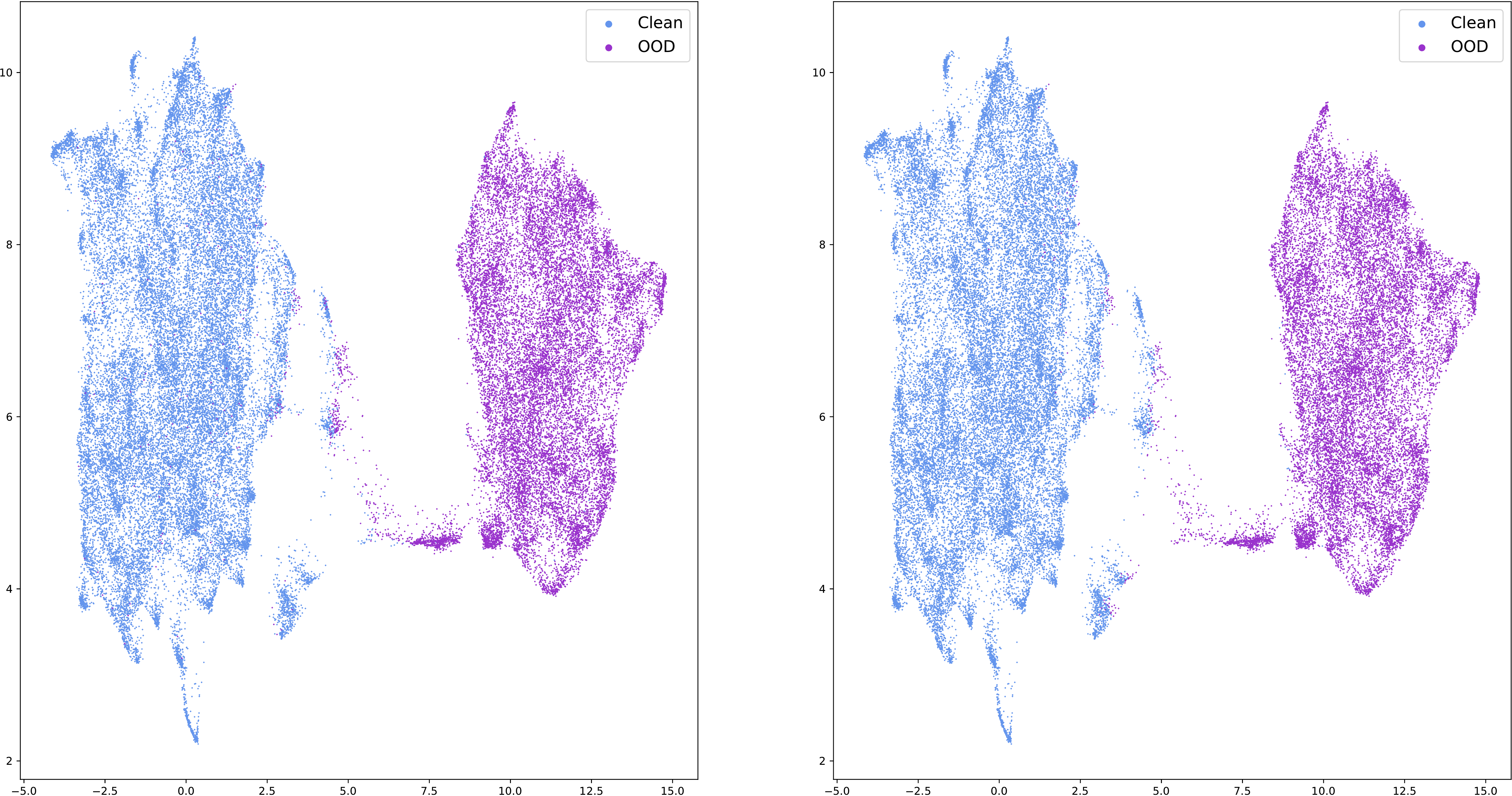"}
\par
\caption{Visualization of the separation of OOD clusters on the CNWL dataset corrupted with $40\%$ web noise. The right hand plot shows the decision made by the 2D Gaussian Mixture when fit to the embedded features $E$ while the left hand plot is the ground-truth. \label{fig:CNWL}}
\end{figure*}

We observe that the samples can be clustered at the dataset level on the CNWL dataset~\cite{2020_ICML_MentorMix} where little ID noise is present. Figure~\ref{fig:CNWL} is a UMAP~\cite{2018_arXiv_UMAP} visualization of the separation of the features when the web noise is injected at $40\%$ and a visual appreciation of how well the noise is captured by the 2D Gaussian mixture.
\begin{figure*}
\centering
\includegraphics[width=.9\linewidth]{"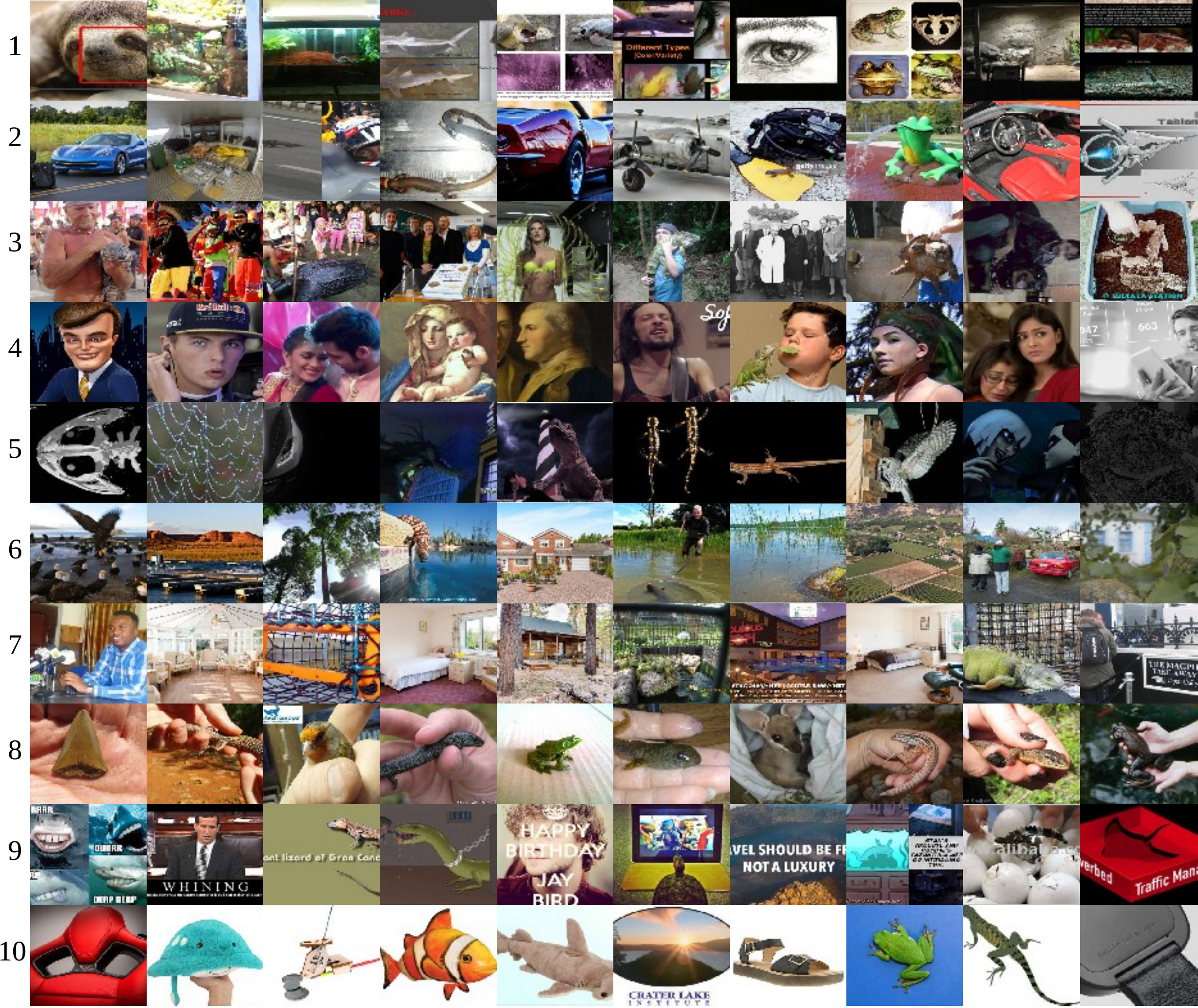"}
\par
\caption{Visualization of OOD clusters captured from the embedding. \label{fig:webclust}}
\end{figure*}
\section{Detected clusters of OOD samples on Webvision}
Figure~\ref{fig:webclust} shows examples of images from captured OOD clusters on Webvision. Some seem to capture very basic features such as black backgrounds in cluster $5$, white backgrounds in cluster $10$, or a basic square shape in cluster $1$. More complex features are also captured such as human bodies in cluster $3$, faces in cluster $4$, hands in cluster $8$, or text in cluster $9$. Including these images when training the network helps improve low-level features and the accuracy when predicting on ID data (see the ablation table in the main paper).

\section{Fast noise annotation using the linear separability of OOD samples}
Although we propose to use an automatic algorithm to detect clusters in the ordered chain computed in OPTICS, we believe that for real world applications, this task could easily (and probably more accurately) be done by a human annotator, circling the clusters and selecting a few random images from each cluster to classify between clean or OOD at the cluster level. This would not be a lengthy task and would be favorable to real world applications. 

\section{OPTICS hyper-parameters}
Table~\ref{tab:hyperclust} references the hyper-parameters used for detecting the OOD noise cluster and outliers, where $\xi$ is the only sensitive hyper-parameter of the cluster detection algorithm in OPTICS that we provide a study for in the main body of the paper. We scan the embedded features at three neighborhood sizes and select the neighborhood leading to the lowest amount of outliers. For the CNWL dataset, we find that using a spherical covariance in the Gaussian mixture is better when the number of samples in the clean and OOD set is unbalanced, i.e.~for $20\%$ and $80\%$ of noise, a spherical covariance captures the clusters more accurately. We use the full covariance setting otherwise.

\begin{table}[t]
    \caption{hyper-parameters for clustering the embedded contrastive features $E$ using OPTICS. \label{tab:hyperclust}}
    \global\long\def\arraystretch{0.9}%
    \centering
    \resizebox{.6\textwidth}{!}{%
    \centering
    \begin{tabular}{l>{\centering}c>{\centering}c>{\centering}c>{\centering}c>{\centering}c>{\centering}c>{\centering}c>{\centering}c>{\centering}c>{\centering}c}
    \toprule
    Dataset & corruption & resolution & neigh. size & min size & $\xi$  \tabularnewline
    \midrule
    \multirow{2}{*}{CIFAR-100} & INet & $32\times32$ & $\{75, 50, 25\}$ & $75$ & $0.01$  \tabularnewline
     & Places & $32\times32$ & $\{75, 50, 25\}$ & $75$ & $0.01$ \tabularnewline
    \midrule
    Webvision & Web & $84\times84$ & $\{75, 50, 25\}$ & $50$ & $0.01$  \tabularnewline
    \bottomrule
    \end{tabular}}
\end{table}

\section{SCNF algorithm}
Algorithm~\ref{alg:algo} presents pseudocode for the SNCF algorithm.
\begin{algorithm*}[]
\caption{SNCF \label{alg:algo}}
\textbf{Input}: $\mathcal{D} = \left\{ \left(x_{i},y_{i}\right)\right\} _{i=1}^{N} $ a web noise dataset. $h$ a randomly initialized CNN. $g$ a CNN pretrained using self-supervised constrastive learning on $\mathcal{D}$\\
\textbf{Parameters}: $\alpha, e_\text{warmup}, e_\text{max}, \xi, \tau$, neigh\\
\textbf{Output}: Trained neural network $h$
\begin{algorithmic}[1] 
\State feats $ = g(D)$ \Comment{Extract unsup. contrastive features}
\State embed = Embedding(feats, neigh) 
\For{$c = 1, \dots C$} \Comment{Apply OPTICS per class}
\State embedC = embed[class == c]
\State clusterClean, clusterOod, outliers = OPTICS(embedC, $\xi$)
\State allClean, allOod, allIdn $\xleftarrow{}$ clusterClean, clusterOod, outliers
\EndFor
\State
\State embedOod = Embedding(feats[allOod], neigh) \Comment{Re-embed without ID}
\State simsOod = OPTICS(embedOod, $\xi$) \Comment{Discover similar OOD clusters}
\State
\For{$e = 1, \dots e_\text{warmup}$} \Comment{Warmup}
\For{$t=1, \dots \mathit{numBatches}$}
\State Sample the next mini-batch $(x,y)$ from $\mathcal{D}[all_\text{clean}]$
\State $l = \mathrm{CrossEntropy}(h(x_\text{mixed}), y_\text{mixed})$
\State $h = \mathrm{UpdateNetworkWeights}($L$)$
\EndFor
\EndFor
\State
\For{$e = e_\text{warmup}+1, \dots e_\text{max}$} \Comment{Noise robust training}
\For{$t=1, \dots \mathit{numBatches}$}

\State Two weakly augmented views $(x,y)$ and $(x',y)$ from $\mathcal{D}[\mathit{allClean}]$
\State Two weakly augmented views $(z,)$ and $(z',)$ from $\mathcal{D}[\mathit{allIdn}]$
\State $w =$ ConsistencyReg($h(z)$, $h(z')$, $\tau$) \Comment{Guessing labels for idn images}
\State Supervised mini-batch $X = (x, x', z)$ with labels $Y = (y,y,w)$
\State $X_\text{mix}, Y_\text{mix}$ = mixup$(X, Y, \alpha)$ \Comment{Mixup}
\State $l_\text{ce}$ = CrossEntropy$(h(X_\text{mix}), Y_\text{mix})$
\State
\State Weak augs $Q = (x, z, o)$ from $\mathcal{D}[\mathit{allClean}], \mathcal{D}[\mathit{allIdn}], \mathcal{D}[\mathit{allOod}]$
\State Strong augs $Q' = (x'', z'', o'')$ from $\mathcal{D}[\mathit{allClean}], \mathcal{D}[\mathit{allIdn}], \mathcal{D}[\mathit{allOod}]$
\State sims = ComputeSims($y$, $w$, simsOod) 
\State $l_\text{cont} = $ GuidedContLoss($h(Q)$, $h(Q')$, sims) \Comment{Cont feats through proj.}
\State
\State $h = \mathrm{UpdateNetworkWeights}(l_\text{ce}+l_\text{cont})$
\EndFor
\EndFor
\State \textbf{return} $h$ \Comment{Robustly trained network}
\end{algorithmic}
\end{algorithm*}

%
%
\newpage
\bibliographystyle{splncs04}
\bibliography{egbib}